\documentclass{article}
\usepackage{spconf,amsmath,amssymb,graphicx,booktabs,array}
\usepackage[T1]{fontenc}
\usepackage[utf8]{inputenc}
\usepackage{microtype}
\usepackage[hidelinks]{hyperref}
\newcommand{\gainGestureBridge}{12.31}
\newcommand{\forceRelativeGain}{10.92}
\title{EMGBlend: Heterogeneity-Aware Self-Supervised Pretraining\\for Gesture and Force Decoding}

\name{Yuwei Jia$^{1,3}$ \quad Cheng Zhong$^{2,3}$ \quad Jinyang Yu$^{3}$ \quad Zhe Cui$^{1,*}$}
\address{
$^{1}$Beijing University of Posts and Telecommunications, China\\
$^{2}$Shenzhen University, China;\quad $^{3}$Dexwise, China}
\begin{document}
\ninept
\maketitle

\begin{abstract}
Public surface electromyography (EMG) datasets vary widely in electrode
layout, channel count, frequency support, and size. Simply mixing them for
pretraining can misalign channel semantics, introduce spectral targets that
some devices cannot observe, and let large or high-channel-count datasets
dominate learning. We introduce EMGBlend, a self-supervised framework designed
around these differences. It combines shared channel patches with
geometry-aware attention, restricts spectral targets to each recording's
supported frequency band, and balances exposure across data sources.
We pretrain a 109M-parameter model on 11 public EMG sources and evaluate it on
gesture recognition, continuous-force regression, and contact classification.
EMGBlend consistently outperforms matched random initialization and waveform
reconstruction controls. Fixed-budget source controls show that multi-source pretraining improves
  gesture recognition and remains competitive for force decoding. Ablations confirm that
geometry, band-aware targets, and source balancing each contribute to transfer,
although cross-person NinaPro force estimation remains difficult. Overall,
EMGBlend shows how heterogeneous EMG datasets can be combined through explicit
mechanism design rather than simple concatenation. Code is available at \url{https://github.com/tamanano/EMGBlend}
\end{abstract}


\section{Introduction}
Surface electromyography (EMG) captures movement and exerted force.
Public datasets cover hand motions, typing, activities, and high-density signals
\cite{salter2024emg2pose,sivakumar2024emg2qwerty,kaifosh2025generic,delpreto2022actionsense,jiang2021hyser},
but differ in arrays, sampling rates, passbands, and annotations.
Pooling these recordings could enlarge the training base for reusable
EMG representations. However, naive pooling conflates sensor identities,
assigns targets outside some devices' observable bands, and overexposes
sources with more windows or channels. BIOT provides a channel-segment interface for cross-dataset biosignals
\cite{yang2023biot}; PhysioWave uses wavelet features and frequency-guided
masking~\cite{chen2025physiowave}; and EMBridge uses paired EMG and pose
to guide gesture transfer~\cite{cui2026embridge}. Recent works
explore shared-channel multi-task encoders
  \cite{fasulo2025tinymyo}, spectral pseudo-labels for movement decoding
  \cite{weng2025spectre}, and quantized biosignal representations
  \cite{barmpas2025neurorvq,avramidis2025biocodec}. These approaches
provide relevant interface, target, and transfer ingredients, but do not
jointly address montage, frequency-support, and source-exposure mismatch.
Our contribution is the coupled design of these mechanisms for heterogeneous
EMG pooling, rather than dataset concatenation alone. We include prior systems
as contextual references and use matched internal controls for the main
mechanistic comparisons.

EMGBlend therefore treats heterogeneous pooling as a coupled interface,
target, and exposure problem.
First, shared channel patches and geometry-aware temporal--spatial
attention provide a common interface for variable sensor layouts.
Second, supported-band spectral codes avoid assigning targets to
frequencies absent from a recording. Third, source-exposure correction
tempers imbalance induced jointly by corpus size and channel count.
These mechanisms require neither shared task labels nor paired pose.

We evaluate the resulting initialization through pose-guided gesture transfer
(Table~\ref{tab:gesture}), frozen gesture recognition, continuous-force and
contact decoding, matched waveform objectives, and one-mechanism-at-a-time
removals. Fixed-budget source controls separate source diversity from sample
count, while a capacity sweep tests whether gains follow model size alone.
With the same capacity, updates, and number of windows, eleven-source
pretraining improves all four DB7 settings over a single source and remains
comparable on PiMForce. A common-protocol comparison further evaluates an
external architecture under matched downstream splits and metrics.
EMGBench's generalization/adaptation distinction~\cite{yang2024emgbench}
motivates explicit user splits and supervision (Sec.~\ref{sec:settings}).

\section{Heterogeneity-Aware EMG Pretraining}
\label{sec:method}
\begin{figure*}[t]
\centering
\includegraphics[width=0.9\textwidth]{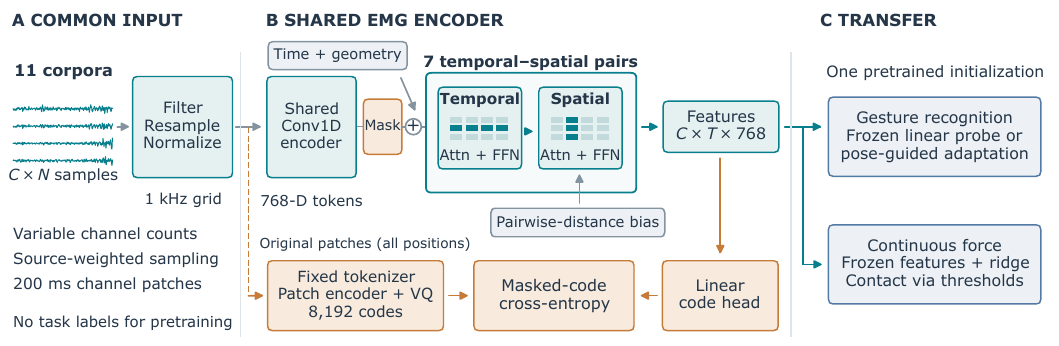}
\caption{EMGBlend's heterogeneity-aware architecture. Shared Conv1D patch
embeddings feed seven temporal--spatial attention pairs with sensor geometry.
The frozen tokenizer supplies code targets (orange); encoder features
support downstream decoding (blue). Attention grids indicate axes,
not measured weights.}
\label{fig:method}
\vspace{-10pt}
\end{figure*}

We pool sources $\mathcal D=\bigcup_{s=1}^{S}\mathcal D_s$, where recording
$X_s\in\mathbb R^{C_s\times N_s}$ has source-dependent channel count,
sampling rate, frequency support, and geometry $G_s$
(Fig.~\ref{fig:method}). The framework couples three corresponding
mechanisms: a montage-flexible interface, recording-valid targets, and
source-exposure correction. Task labels are used only downstream.

\subsection{Source-aware pooling across acquisition settings}
Recordings are filtered within their usable band, resampled to 1\,kHz,
and normalized per channel by the full recording's median and interquartile
range, including at test time as offline preprocessing. Each channel is divided
into non-overlapping 200-sample patches. Available coordinates and
angles accompany the signal; unknown geometry uses learned embeddings.
Resampling standardizes timing without restoring absent frequencies.

To reduce domination by large or high-channel-count corpora, the loader
tempers channel-patch exposure imbalance. If $n_s$ is the sum of capped
channel counts across the available windows of source $s$, a candidate
window is retained with probability
\begin{equation}
 a_s(\alpha)=\left(\frac{\min_r n_r}{n_s}\right)^{\alpha}.
\end{equation}
The standard recipe uses $\alpha=1/2$; the $\alpha=0$ ablation gives
$a_s(0)=1$ for every source and therefore accepts every candidate window.
Channel-count bucketing and a nominal channel budget limit padding;
at most 64 channels are sampled per window. Available-pool size and
realized training exposure remain distinct quantities.

\subsection{Geometry-aware variable-montage encoder}
A shared Conv1D patch encoder maps channels to tokens, following the
channel-segment interface of BIOT~\cite{yang2023biot}. Geometry uses
coordinate Fourier features, six angular harmonics, and a 16-bin
pairwise-distance attention bias; missing geometry uses learned embeddings.
Temporal and geometry embeddings precede alternating within-channel temporal
and across-channel spatial attention; invalid channels are masked. Following divided
attention~\cite{bertasius2021timesformer}, attention computation scales as
$O(CT^2+TC^2)$ at fixed width for $C$ channels and $T$ patches. The
representation does not require a shared device or montage identity: geometry
describes sensor arrangement without assuming anatomical registration.

\subsection{Supported-band spectral-code learning}
\begin{figure}[t]
\centering
\includegraphics[width=0.9\columnwidth]{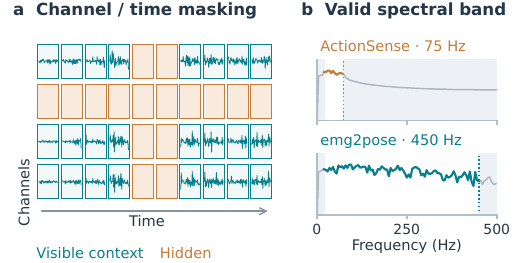}
\caption{Supported-band target construction. (a) Real emg2pose training signals with
an illustrative whole-channel/time mask; hidden patches are orange.
(b) Log-amplitude spectra from ActionSense and emg2pose training
patches on the same 1\,kHz grid. Gray regions are excluded from the
spectral reconstruction loss. Cutoffs reflect these recordings.}
\label{fig:maskband}
\vspace{-10pt}
\end{figure}

We follow masked representation learning~\cite{he2022mae} with discrete
spectral targets, drawing on vector quantization~\cite{oord2017vqvae}
and LaBraM's neural-spectrum prediction for EEG~\cite{jiang2024labram}.
A vector-quantized tokenizer supplies one code per channel patch.
Its decoder learns the
log-amplitude spectrum of a mean-centered patch, with reconstruction
restricted to frequencies supported by that recording. For spectral
bins $k$ and valid-band mask $b_{s,k}$, the reconstruction term is
\begin{equation}
 \mathcal L_{\mathrm{spec}}=
 \mathbb E_{x\sim\mathcal D_s}
 \frac{\sum_k b_{s,k}(\widehat a_k-a_k)^2}{\sum_k b_{s,k}},
\end{equation}
where $a_k=\log(|\mathcal F(x-\bar x)_k|+10^{-6})$.
Averaging over valid bins avoids downweighting narrower passbands
for having fewer bins. Figure~\ref{fig:maskband} illustrates real inputs.

The tokenizer has 8,192 EMA codes and is trained for 20k updates with 32
windows (at most 16,384 patches) per update using AdamW (learning rate
$3\times10^{-4}$, weight decay 0.01), then fixed during backbone training.
Targets use original patches, including positions later hidden.
Whole-channel and temporal masking hide raw-patch embeddings; the
encoder predicts their discrete codes from the remaining context:
\begin{equation}
 \mathcal L_{\mathrm{SSL}}=-\mathbb E_{X\sim\mathcal D}
 \frac{1}{|\mathcal M|}\sum_{(c,t)\in\mathcal M}
 \log p_\theta(z_{c,t}\mid\widetilde X,G).
\end{equation}
Here $\mathcal M$ contains hidden, valid positions. Both masking fractions
are 0.25. The common code-prediction objective permits joint training without
reconciling source gesture vocabularies or regression labels.

\subsection{Transfer to gestures and force}
Frozen gestures use regularized linear classifiers on pooled features.
For adapted gestures, EMGBlend initializes EMBridge's EMG encoder while
retaining query alignment, pose reconstruction, and pose-side assets
\cite{cui2026embridge}. The encoder is frozen after adaptation for
probing; ``SSL'' denotes pretraining, while adaptation uses pose labels.
Retrieval additionally uses a pose library.
For force, ridge regression maps frozen features to finger targets;
PiMForce contact decisions threshold predictions using validation users.
Matched random encoders control feature dimension.

\section{Experiments}
\subsection{Training and evaluation settings}
\label{sec:settings}
\begin{figure}[t]
\centering
\includegraphics[width=\columnwidth]{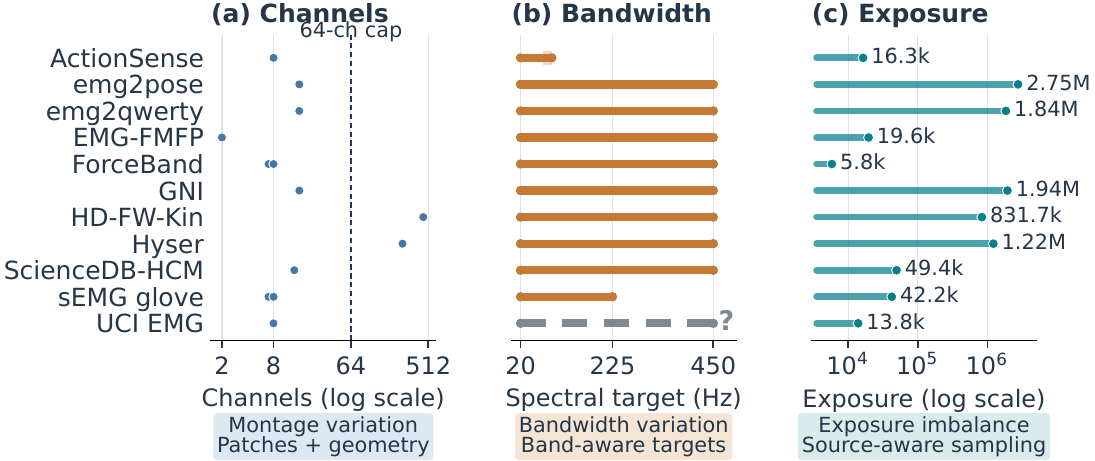}
\caption{Heterogeneity of the 11-source pretraining pool. (a) Released
channel counts and the 64-channel sampling cap. (b) Recording-supported
frequency ranges used by the spectral target; UCI's effective upper cutoff is
uncertain. (c) Naive available-window $\times$ capped-channel exposure before
source-aware acceptance. The three panels motivate the montage, target, and
sampling mechanisms.}
\label{fig:landscape}
\vspace{-10pt}
\end{figure}

 Unless identified as a control or ablation, Full or SSL denotes the
  109M checkpoint with width 768, seven temporal--spatial pairs, and 12 heads.
  The tokenizer is trained separately and frozen before backbone pretraining.
The 462,980-window pool comprises ActionSense~\cite{delpreto2022actionsense},
emg2pose~\cite{salter2024emg2pose}, emg2qwerty~\cite{sivakumar2024emg2qwerty},
EMG-FMFP~\cite{kosteley2026fmfp}, ForceBand~\cite{he2026forceband},
and GNI~\cite{kaifosh2025generic}. It also includes HD-FW-Kin~\cite{guo2025hdfwkin},
Hyser~\cite{jiang2021hyser}, KIMHu (ScienceDB HCM)~\cite{hernandez2023kimhu},
multimodal glove~\cite{kyranou2025armtranslation}, and UCI EMG~\cite{krilova2018uciemg}.
Across the released sources, channel count ranges from 2 to 448 (capped at 64
when sampled), recording rate from approximately 157 to 2,048 Hz, supported
upper cutoff from approximately 71 to 450 Hz, and available windows per source
from 752 to 171,828. Figure~\ref{fig:landscape} summarizes the resulting montage, 
frequency-support, and exposure heterogeneity and links each source of variation 
to the corresponding EMGBlend mechanism. The final manifest contains 16,344 recordings.
Tokenizer and backbone training both exclude NinaPro, PiMForce,
emg2pose's official non-training recordings, and the declared Hyser
holdout. The tokenizer is trained on the allowed eleven-source pool
and then frozen to supply SSL targets; it is not used in downstream
inference. Exclusions are defined by recording identity before sampling, and
the consumption audit found no excluded recordings. UCI's nominal 1-kHz
timestamp grid contains repeated rows,
so we do not infer its physical acquisition bandwidth from that grid.
Because UCI downstream subjects occurred in unlabeled pretraining, we omit UCI
from transfer comparisons rather than claim subject-disjoint generalization.
The 9M--109M models share 60k AdamW updates, 3k warmup steps,
cosine decay, weight decay 0.05, and peak learning rate
$1.96\times10^{-3}$. Each of two workers has a nominal 2,048-channel
budget; bucketing makes window/token counts variable. Under memory and
stability constraints, 267M instead uses four workers, a 768-channel budget
per worker, 6k warmup, and base/effective learning rates
$2.5\times10^{-4}/7.07\times10^{-4}$.
Our local splits differ from EMGBench's
leave-one-subject-out and chronological adaptation benchmarks:
pose adaptation uses training users, without test-user calibration.

\textbf{Gesture protocols.}
emg2pose has four seen and four unseen gesture-stage groups. Unseen
groups are excluded from pose adaptation but labeled for probing:
12 probe users (3,012 windows) and 20 test users (6,212 windows) are
disjoint. Adaptation uses 40 epochs, batch size 256, and learning rate
$4\times10^{-4}$. Linear probes use 300 AdamW steps; the SSL row in
Table~\ref{tab:gesture} reports the mean over seeds 7 and 8.
Frozen layer/pooling choices use grouped validation within probe users.
On NinaPro DB7~\cite{krasoulis2017db7}, B3/C3 use exercise B/C local IDs
$\{1,5,10\}$, while B4/C4 add ID 15. Frozen final-layer mean/std features use
ten outer splits with disjoint train/test users; participants may recur
across splits. Feature scaling and
regularization use nested person-level validation. These are defined
3/4-class subsets rather than the full DB7 vocabulary.

\textbf{Force protocols.}
NinaPro DB2/DB3 exercise 3 (E3)~\cite{atzori2014ninapro} uses two-second windows and
six window-mean force targets. Within-person testing trains on
repetitions 1/3/4/6 and tests on 2/5, over 40 DB2 and 11 DB3 participants.
Cross-person DB2 uses five subject folds and grouped inner validation.
Feature and target scaling use training rows only, including within
each inner fold. Validation holds out training repetitions or users.
Predictions are inverted to native sensor units for MAE and standard
variance-weighted $R^2$, averaged over users. No test-label statistics
enter prediction; conversion of native units to newtons is unverified.
Handcrafted controls use root-mean-square (RMS), mean absolute value,
and waveform length.

PiMForce~\cite{seo2024pimforce} tests five continuous fingertip targets
across 21 users in five folds. Training/validation use S1--S2 of non-test
users; testing uses S3 of held-out users (21,054 windows). Readouts use
training-user weighting and validation-selected ridge regularization.
MAE is in calibrated target units. Contact labels use a force-sensitive
resistor (FSR) threshold of $1.0$;
prediction thresholds are selected on validation users. Preprocessing
is offline. Repeats and uncertainty retain their probe or participant
unit; they are not independent pretraining repetitions.

\begin{table}[t]
\centering
\caption{emg2pose gesture BA (\%). LP: linear probe; Ret.: pose retrieval.
EMGBlend averages adaptation seeds 7/8; random and local
EMBridge use single runs. Published rows are contextual references.}
\label{tab:gesture}
\vspace{3pt}
\small\setlength{\tabcolsep}{1.5pt}
\begin{tabular*}{\columnwidth}{@{\extracolsep{\fill}}lrrrr@{}}
\toprule
 & \multicolumn{2}{c}{Seen} & \multicolumn{2}{c}{Unseen} \\
\cmidrule(lr){2-3}\cmidrule(l){4-5}
Method & LP & Ret. & LP & Ret. \\
\midrule
\textbf{EMGBlend SSL (ours)} & \textbf{81.76} & \textbf{79.70} & \textbf{61.68} & \textbf{55.99} \\
EMGBlend 109M, random init. & 68.96 & 65.51 & 48.48 & 43.88 \\
EMBridge (reprod.)~\cite{cui2026embridge} & 76.62 & 74.80 & 49.37 & 47.18 \\
\midrule
EMBridge~\cite{cui2026embridge} & 78.50 & 77.70 & 50.50 & 52.80 \\
emg2pose~\cite{salter2024emg2pose} & 73.40 & -- & 40.50 & -- \\
\bottomrule
\end{tabular*}
\vspace{-10pt}
\end{table}

\begin{table}[t]
\centering
\caption{Frozen 109M force decoding. HC: handcrafted features. NinaPro MAE
uses native sensor units; PiMForce MAE uses calibrated FSR units.}
\label{tab:force}
\vspace{3pt}
\setlength{\tabcolsep}{3.5pt}
\begin{tabular}{lrrr}
\toprule
Dataset / metric & HC & Random & SSL \\
\midrule
DB2 within / $R^2$ & 0.7863 & 0.3423 & \textbf{0.8562} \\
DB2 within / MAE & 0.9734 & 1.8705 & \textbf{0.8704} \\
DB3 within / $R^2$ & 0.2781 & 0.0887 & \textbf{0.4932} \\
DB3 within / MAE & 1.7726 & 2.0220 & \textbf{1.4662} \\
DB2 cross / $R^2$ & \textbf{-0.0363} & -0.2218 & -0.2926 \\
DB2 cross / MAE & \textbf{2.1314} & 2.3287 & 2.3388 \\
PiMForce / MAE & 2.6151 & 2.3802 & \textbf{2.1204} \\
PiMForce / BA (\%) & 73.94 & 79.75 & \textbf{82.31} \\
PiMForce / F1 (\%) & 53.58 & 60.69 & \textbf{65.60} \\
\bottomrule
\end{tabular}
\vspace{-10pt}
\end{table}

\begin{table*}[t]
\centering
\caption{System-level comparison with the PhysioWave
architecture~\cite{chen2025physiowave} under common downstream splits and
metrics. Pretraining setup and model capacity are not matched. For DB5 and PiMForce, we use the officially released PhysioWave model. Because this model was trained on EPN-612, the EPN-612 entry uses a retrained
PhysioWave architecture. UCI is omitted because its downstream subjects were
not held out from EMGBlend pretraining.}
\label{tab:external}
\small
\setlength{\tabcolsep}{1.8pt}
\renewcommand{\arraystretch}{0.95}
\begin{tabular*}{\textwidth}{@{\extracolsep{\fill}}lr*{10}{r}@{}}
\toprule
& & \multicolumn{3}{c}{EPN-612} & \multicolumn{4}{c}{DB5} & \multicolumn{3}{c}{PiMForce} \\
\cmidrule(lr){3-5}\cmidrule(lr){6-9}\cmidrule(l){10-12}
Method & Params & Acc & BA & mF1 & Acc & BA & wF1 & mF1 & MAE & BA & F1 \\
\midrule
PhysioWave arch. & 4.99M & 90.73 & 90.73 & 90.77 & 77.16 & 56.11 & 76.40 & 55.40 & 2.2993 & 81.17 & 61.90 \\
\textbf{EMGBlend} & 109M & \textbf{92.78} & \textbf{92.78} & \textbf{92.80} & \textbf{95.29} & \textbf{91.03} & \textbf{95.26} & \textbf{91.16} & \textbf{2.1204} & \textbf{82.31} & \textbf{65.60} \\
\bottomrule
\end{tabular*}
\par\vspace{1pt}\noindent

\vspace{-10pt}

\end{table*}

\begin{table*}[t]
  \centering
  \caption{Unified objective, mechanism, source-pool, and capacity controls
  under common frozen downstream protocols. DB7 entries report ten-split
  means; emg2pose entries average five linear-probe initializations; and
  DB2/DB3 report participant-averaged within-subject $R^2$. Best values are
  bold and second-best are underlined.}
  \label{tab:unified}
  \label{tab:source109}
  \label{tab:components}
  \label{tab:scale}
  \small
  \setlength{\tabcolsep}{0.45pt}
  \renewcommand{\arraystretch}{0.92}
  \begin{tabular*}{\textwidth}{@{\extracolsep{\fill}}lrrr*{11}{r}@{}}
  \toprule
  & & & & \multicolumn{2}{c}{emg2pose BA $\uparrow$} & \multicolumn{4}{c}{DB7 BA $\uparrow$} & \multicolumn{2}{c}{NinaPro $R^2\uparrow$} & \multicolumn{3}{c}{PiMForce} \\
  \cmidrule(lr){5-6}\cmidrule(lr){7-10}\cmidrule(lr){11-12}\cmidrule(l){13-15}
  Variant & Param. & Src. & Win. & Seen & Unseen & C3 & B4 & B3 & C4 & DB2 & DB3 & MAE $\downarrow$ & BA $\uparrow$ & F1 $\uparrow$ \\
  \midrule
  \multicolumn{15}{l}{\textit{Objective controls}} \\
  Waveform MAE & 109M & 11 & 463k & 54.14 & 37.74 & 38.04 & 49.56 & 63.08 & 42.90 & 0.3150 & 0.1062 & 2.3559 & 79.73 & 61.41 \\
  Generic masked MAE & 109M & 11 & 463k & 51.09 & 38.67 & 38.56 & 46.63 & 58.54 & 43.36 & 0.3178 & 0.0871 & 2.3136 & 79.44 & 61.21 \\
  \midrule
  \multicolumn{15}{l}{\textit{Mechanism removals}} \\
  $\alpha=0$ & 109M & 11 & 463k & 55.74 & 38.15 & 48.77 & 62.61 & 74.38 & 54.13 & 0.7929 & 0.4326 & 2.2852 & 79.43 & 61.66 \\
  No geometry & 109M & 11 & 463k & 39.37 & 33.26 & 48.22 & 53.09 & 63.23 & 51.69 & 0.5495 & 0.2576 & 2.5441 & 75.38 & 55.39 \\
  All spectral bins & 109M & 11 & 463k & 53.13 & 41.81 & 48.95 & 55.63 & 66.20 & 52.47 & 0.7120 & 0.3474 & 2.2943 & 79.35 & 60.39 \\
  \midrule
  \multicolumn{15}{l}{\textit{Source-pool controls}} \\
  Random initialization & 109M & 0 & -- & 49.38 & 36.19 & 36.48 & 44.78 & 58.28 & 41.65 & 0.3423 & 0.0887 & 2.3802 & 79.75 & 60.69 \\
  Single$^{\dagger}$ & 109M & 1 & 172k & \textit{74.23} & \textit{49.11} & 48.64 & 64.87 & 72.84 & 53.00 & \textbf{0.8598} & 0.4605 & 2.1743 & 81.56 & 64.79 \\
  Nine & 109M & 9 & 172k & 42.41 & 34.41 & 45.60 & 57.74 & 69.07 & 50.17 & 0.5992 & 0.2881 & 2.3274 & 78.79 & 59.49 \\
  Eleven-matched & 109M & 11 & 172k & 68.29 & 47.55 & \textbf{52.75} & \underline{65.60} & \textbf{74.53} & \underline{56.29} & 0.8413 & 0.4529 & 2.1633 & 81.60 & 64.42 \\
  \midrule
  \multicolumn{15}{l}{\textit{Capacity controls (eleven-source full pool)}} \\
  9M & 9M & 11 & 463k & 66.71 & \underline{47.86} & 46.04 & 61.43 & 69.29 & 53.26 & 0.8042 & 0.4132 & 2.2487 & 80.99 & 63.68 \\
  12M & 12M & 11 & 463k & \underline{68.77} & \textbf{48.55} & 45.74 & 63.57 & 72.75 & 53.26 & 0.8084 & 0.4261 & 2.2663 & 80.55 & 63.67 \\
  25M & 25M & 11 & 463k & 67.96 & 45.46 & 49.05 & 63.43 & 73.54 & 53.96 & 0.8204 & 0.4440 & 2.1870 & 81.97 & 65.01 \\
  50M & 50M & 11 & 463k & 68.02 & 46.34 & 48.72 & 65.12 & 74.24 & 54.29 & 0.8401 & \underline{0.4719} & \underline{2.1467} & \underline{82.13} & \textbf{65.62} \\
  267M$^{*}$ & 267M & 11 & 463k & \textit{69.07} & \textit{48.63} & \textit{51.64} & \textit{66.57} & \textit{75.68} & \textit{56.50} & \textit{0.8610} & \textit{0.4930} & \textit{2.1516} & \textit{81.65} & \textit{64.94} \\
  \midrule
  \textbf{Full method} & 109M & 11 & 463k & \textbf{70.39} & 46.57 & \underline{50.63} & \textbf{65.75} & \underline{74.47} & \textbf{56.38} & \underline{0.8561} & \textbf{0.4773} & \textbf{2.1204} & \textbf{82.31} & \underline{65.60} \\
  \bottomrule
  \end{tabular*}
  \par\vspace{1pt}\noindent
  \begin{minipage}{\textwidth}
  \small
  $^{\dagger}$Single-source pretraining may overfit to emg2pose; its italicized
  emg2pose results are excluded from ranking. $^{*}$The 267M run uses a
  smaller batch because of GPU memory limits, so its italicized results are
  reported but excluded from controlled comparisons.
  \end{minipage}
\vspace{-10pt}

\end{table*}

\subsection{Gesture recognition and transfer}
After pose-guided adaptation (Table~\ref{tab:gesture}), EMGBlend
reaches $81.76\pm0.37\%$ seen LP and $61.68\pm0.30\%$ unseen LP;
the local EMBridge reproduction reaches 76.62\% and 49.37\%, respectively.
EMGBlend improves all four endpoints, including \gainGestureBridge{} percentage
points (pp) on unseen LP. Since the official code, checkpoint, and exact data
assignments are unavailable, this is a clean-room comparison under our
documented protocol rather than an exact benchmark reproduction.
The matched 109M random initialization reaches 68.96\% seen LP and
48.48\% unseen LP, giving SSL gains of 12.80 and 13.20 pp.

Frozen controls in Table~\ref{tab:unified} show the same pattern: Full
  improves both emg2pose groups and all evaluated gesture settings over matched
  random initialization while holding architecture fixed.

\subsection{Continuous force decoding}
Within-person DB2/DB3 favor SSL over both controls.
Cross-person DB2 remains difficult: the SSL $R^2$ is $-0.2926$,
below both matched random ($-0.2218$) and handcrafted ($-0.0363$),
so this endpoint remains a negative result rather than evidence of
cross-person force transfer. On PiMForce, SSL reduces cross-user MAE from random's 2.3802 to
2.1204 (\forceRelativeGain{}\%), versus HC's 2.6151. Contact BA/F1
also improve over matched random features. MAE includes both contact and
non-contact windows, whereas BA/F1 evaluate thresholded contact state.

\subsection{External baselines and unified controls.}
Table~\ref{tab:external} provides a system-level comparison under common
  downstream splits and metrics. Model capacity, pretraining data, and training
  budget are not matched, so the comparison does not isolate SSL design. For DB5,
  we use a shared downstream protocol rather than PhysioWave's original split.

  All objective and mechanism controls use the 109M backbone, eleven-source pool,
  and 60k updates. Waveform MAE replaces the target; Generic masked MAE also
  disables geometry and source correction. The remaining rows remove one
  mechanism at a time, and Full exceeds every such control on all 11 endpoints.

  Source-pool controls compare emg2pose-only, nine-source, count-matched
  eleven-source, and full-pool pretraining. At matched volume, multi-source
  pretraining improves gesture recognition and remains competitive for force
  decoding. Scaling is non-monotonic, so capacity alone does not explain the
  gains; ranking exclusions are listed in Table~\ref{tab:unified}.

\section{Conclusion}
EMGBlend couples montage-flexible inputs, supported-band targets, and
source-balanced sampling for heterogeneous EMG pretraining. Matched controls
support each mechanism, and eleven-source training improves all DB7 settings at
fixed capacity and window budget. Full improves PiMForce, while cross-person
force decoding remains an open challenge.

\newpage
{
\noindent\textbf{Acknowledgments:} OpenAI Codex and Claude Code assisted with
language editing, implementation summaries, LaTeX organization, and code
writing. The authors verified all content and take full responsibility; no
AI-generated data or experimental results were used.

\noindent\textbf{Compliance with Ethical Standards:}
This study analyzes public datasets and involved no new participant recruitment,
interaction, or human-subject data collection.}

{
\bibliographystyle{IEEEbib}
\interlinepenalty=10000
\bibliography{references}}
\end{document}